\documentclass[10pt,journal,compsoc]{IEEEtran}
\usepackage{ctable}
\usepackage{caption}
\usepackage{graphicx}
\usepackage{float}
\usepackage{amssymb,amsfonts,amsmath,amsthm}
\usepackage[hidelinks]{hyperref}

\begin{document}

\title{Offensive Language Detection on Social Media Using XLNet}

\author{Reem~Alothman, Hafida~Benhidour, Said~Kerrache
\IEEEcompsocitemizethanks{\IEEEcompsocthanksitem King Saud University, College of Computer and Information Sciences, Riyadh, 11543, KSA.\protect\\
E-mail: hbenhidour@ksu.edu.sa}
}

\IEEEtitleabstractindextext{
\begin{abstract}
The widespread use of text-based communication on social media—through chats, comments, and microblogs—has improved user interaction but has also led to an increase in offensive content, including hate speech, racism, and other forms of abuse. Due to the enormous volume of user-generated content, manual moderation is impractical, which creates a need for automated systems that can detect offensive language. Deep learning models, particularly those using transfer learning, have demonstrated significant success in understanding natural language through large-scale pretraining. In this study, we propose an automatic offensive language detection model based on XLNet, a generalized autoregressive pretraining method, and compare its performance with BERT (Bidirectional Encoder Representations from Transformers), which is a widely used baseline in natural language processing (NLP). Both models are evaluated using the Offensive Language Identification Dataset (OLID), a benchmark Twitter dataset that includes hierarchical annotations. Our experimental results show that XLNet outperforms BERT in detecting offensive content and in categorizing the types of offenses, while BERT performs slightly better in identifying the targets of the offenses. Additionally, we find that oversampling and undersampling strategies are effective in addressing class imbalance and improving classification performance. These findings highlight the potential of transfer learning and XLNet-based architectures to create robust systems for detecting offensive language on social media platforms.
\end{abstract}

}

\maketitle

\IEEEdisplaynontitleabstractindextext

\IEEEpeerreviewmaketitle

\IEEEraisesectionheading{\section{Introduction}
\label{sec:introduction}}
The proliferation of offensive language on social media platforms has become a pressing concern. The anonymity afforded by these platforms enables users to express harmful, aggressive, or disrespectful views that they may not communicate in real life. Offensive content can take many forms, including profanity, harassment, racism, and personal attacks, contributing to toxic online environments. This issue is particularly alarming among adolescent users, who are more susceptible to the psychological and behavioral consequences of exposure to abusive language \cite{foullanguage}.

Given the massive scale and rapid dynamics of social media content, manual moderation is infeasible. As a result, automatic offensive language detection has emerged as a critical task in natural language processing (NLP), aiming to identify and classify abusive language in user-generated text such as tweets, comments, and posts. Traditional machine learning techniques like Support Vector Machines (SVM), Naïve Bayes (NB), and Decision Trees have been explored for this task, but recent research has shown that deep learning models, especially pre-trained transformers, offer superior performance.

Among these, BERT (Bidirectional Encoder Representations from Transformers) has been widely adopted for various NLP tasks and has shown strong results in offensive language detection, including in the SemEval-2019 shared task on identifying and categorizing offensive language in social media (OffensEval) \cite{zampieri2019semeval}. However, while BERT has become a standard baseline, there has been limited exploration of more recent transformer-based models, such as XLNet, in this context.

XLNet \cite{XLNet} is a generalized autoregressive pretraining method that integrates the strengths of autoregressive models and autoencoding models like BERT. It achieves state-of-the-art results on several NLP benchmarks by capturing bidirectional context while preserving the benefits of autoregressive modeling. Despite its reported advantages, XLNet has not been extensively applied to offensive language detection tasks, and no prior work, to the best of our knowledge, has thoroughly evaluated its effectiveness on the OLID dataset.

This research addresses this gap by proposing an offensive language detection approach based on XLNet and evaluating its performance on the OLID dataset across all three subtasks: detecting offensive language, categorizing the offense type, and identifying the offense target. BERT is used as a baseline to benchmark XLNet's performance. The study aims to assess whether XLNet's improved contextual modeling can lead to more accurate and robust offensive language detection, particularly in imbalanced classification scenarios.

The remainder of this paper is organized as follows. Section~\ref{sec:background} provides background on text classification, including traditional and deep learning approaches, word embeddings, and pretrained NLP models. Section~\ref{sec:related-work} reviews prior work on offensive language detection. Section~\ref{sec:design} presents the proposed methodology, including the model architecture and classification pipeline. Section~\ref{sec:experimental} describes the experimental setup, dataset, and evaluation metrics. Finally, Section~\ref{sec:conclusion} concludes the paper and outlines future directions.

\section{Background on Text Classification}
\label{sec:background}
Text classification denotes the task of assigning each text from a collection or a stream of texts to either of the predefined categories. Text classification can be done manually or automatically. It can be used to structure, organize, and categorize the text for different applications. Manual classification is time-consuming and expensive; therefore, many approaches have been proposed for automatic text classification. The text classification problem has a wide variety of applications in various domains of text mining. Here are some examples of domains in which text classification is used: News filtering and Organization, Email Classification and Spam Filtering, and offensive language detection \cite{10.5555/3165154}.

Machine learning approaches perform classifications based on past interpretations, which is a widely used approach in text classification. Using supervised learning (a pre-labeled dataset), a machine learning algorithm can learn the different interpretations between pieces of text and that a specific output is expected for a specific input. The first step towards training a classifier with machine learning is feature extraction. The text is converted into a numerical representation in the form of a vector using different representation techniques. The machine learning algorithm is trained using data consisting of pairs of feature sets (vector representations for each piece of text) and class tags (e.g., politics, sports) to produce a classification model, as shown in \cite{10.5555/3165154}.

When the training is completed on a sufficiently large dataset, the model can make meaningful predictions. In the prediction phase, the feature extractor is again used to create a vector representation of the text that requires a tag to be predicted, as shown in \cite{10.5555/3165154}.

Text classification models commonly employ machine learning techniques such as Support Vector Machines, Bayesian models, and Decision Trees. A wide range of techniques have been developed for text classification. In this section, we will discuss important concepts to help us understand this problem, including text classification approaches, word embeddings, Pre-trained NLP Models, and the evaluation of text classifiers.

\subsection{Early Work on Text Classification}
Early text classification relied on traditional machine learning and rule-based methods. These approaches, though limited in capturing complex semantics, laid the groundwork for modern techniques. This section outlines key classical methods, including Support Vector Machines Classifiers (SVM), Decision Trees, Bayesian classifiers, and rule-based systems.

\subsubsection{Support Vector Machines Classifiers (SVM)}
SVM Classifiers try to partition the data space using linear or non-linear delineations between the different classes. The key in such classifiers is to find the optimal boundaries between the different classes and use them for classification purposes \cite{Classificationbook}.

\subsubsection{Decision Trees (DT)} 
Decision trees are designed using a hierarchical division of the underlying data space, incorporating various text features. The hierarchical division of the data space is intended to create class partitions that are more skewed in terms of their class distribution. For a given text instance, we determine the partition to which it is most likely to belong and use it for classification \cite{Classificationbook}.

\subsubsection{Bayesian Classifiers} 
Bayesian classifiers are also referred to as generative classifiers. They build a probabilistic classifier based on modeling the underlying word features in different classes. The Bayesian idea is then to classify text based on the posterior probability of the documents belonging to different classes, given the presence of words in the documents \cite{Classificationbook}\cite{textclassifcationsurvey}.

\subsubsection{Rule-based Approaches}
There are many ways to categorize rule-based systems based on the following characteristics: input/output values, their length, the type of logic behind them, their structure, and the number of machine learners or others involved.

In rule-based systems, there can be either single or multiple inputs and outputs. From this perspective, these systems can be categorized into four categories based on the length of their inputs and outputs: single input and single output, multiple input and single output, single input and multiple output, and multiple input and multiple output. The features of association rules match all four styles mentioned. For this reason, the rules of association represent the relationship between attributes.
In both the left-hand side (antecedent) and right-hand side (consequent) of the rule, an association rule may have a single or multiple rule term. Therefore, to comply with the uniqueness of the association rules, categorization based on the number of inputs and outputs is essential. There are however, two basic types of association rules: regression rules and classification rules, depending on the type of output value. Regression rules and classification rules may have single or multiple rule terms on the left-hand side, but it is only possible to have one term on the right-hand side. The output values of regression rules must be continuous, while the output values of the classification rules must be discrete.

To create classification rules, there are two approaches: separate and conquer and divide and conquer \cite{10.5555/152181}. A list of if-then rules is generated by the latter method, while the former method produces rules directly in the form of a decision tree. Thus, rule-based systems can be classified into three categories of structure: listed rule-based systems, treed rule-based systems, and networked rule-based systems \cite{RuleBasedSystems, reviewofamodularrulebasedapproach}.

\subsection{Text Representation and Word Embeddings}
Representing text as numerical vectors is a crucial step in enabling machine learning algorithms to process and learn from natural language. Word embeddings are the distributional vector representation of the words introduced to represent their syntax and semantics. Word2vec and GloVe word embedding models have been recently used in various natural language processing applications. In order to deal with various natural language problems and applications, the words in the text must be converted into vectors. Therefore, the semantic similarity between every two words can be measured using standard similarity measures: cosine similarity, Euclidean distance, and others \cite{embeddings}. The most important word embeddings are covered more in-depth.

\subsubsection{Word2vec}
Word2vec is a neural network model that learns word associations from a large corpus of text, representing words as unique lists of numbers or Vectors. After training, it can be used to detect semantically similar words by calculating the cosine similarity of these embedding vectors. Notably, it can utilize two different architectures, as seen in the CBOW (Continuous Bag of Words) model, which attempts to predict the current word based on the words preceding and following it. However, for the Skip-gram model, it is the opposite: surrounding words are predicted using just the current word \cite{embeddings, peters-etal-2018-deep}. 

\subsubsection{GloVe}
Word2Vec is not perfect; it has several disadvantages, including its focus on local context text, which limits its ability to capture the entire context. Due to its flaws, GloVe, or Global Vectors for word representation, was proposed. The naming is chosen because it successfully captures information directly from the global corpus rather than relying on local context windows like the Word2vec model does \cite{embeddings}. The overall results (in terms of accuracy and training time) of this embedding approach in the word analogy task, with comparable experimental settings, can be seen in the. \cite{pennington-etal-2014-glove}

\subsubsection{FastText}
FastText was introduced because the previous method, GloVe, cannot encode unknown words that are not in the vocabulary. This method was introduced by Facebook's AI Research lab and was built upon the idea of Word2Vec, utilizing both the CBOW and Skip-gram models. The main difference between the Word2vec method and fastText is that in fastText, it was proposed to break up words into sub-strings or n-grams and only then input them into the neural network model. Facebook also made 294 language pre-trained models available that are free to use \cite{embeddings}. 

\subsubsection{ELMo}
ELMo, which stands for Embedding from Language Model, is a more advanced word embedding approach that utilizes a 2-layer bi-directional Long Short-Term Memory (LSTM) neural architecture and provides contextualized word embeddings. This can be achieved by utilizing concepts from language models. ELMo word embeddings can be used with NLP systems to achieve higher accuracy, as they consider the context of the words \cite{embeddings}. 

\subsection{Deep Learning Models}
Today, we are in the era of machine learning, as deep learning represents a new paradigm within this field. Deep learning constructs computational models that utilize multiple layers to represent the abstractions of data. Some key enablers of deep learning algorithms, such as generative adversarial networks, convolutional neural networks, and model transfer, have significantly altered our understanding of information processing. Deep learning, which has its roots in conventional neural networks, significantly outperforms its predecessors. It utilizes graph technologies with transformations among neurons to develop many layered learning models. Many of the latest deep learning techniques have been presented and have demonstrated promising results across different kinds of applications such as Natural Language Processing (NLP), visual data processing, speech and audio processing, and many other well-known applications deep learning algorithms perform feature extraction in an automated way, which allows researchers to extract discriminative features with minimal domain knowledge and human effort. These algorithms feature a layered architecture of data representation, where high-level features can be extracted from the network's last layers. In contrast, low-level features are extracted from the lower layers. These kinds of architectures were initially inspired by Artificial Intelligence (AI), simulating the process in key sensorial areas of the human brain \cite{10.1145/3234150}.

\subsubsection{Feed forward Neural networks}
Feed-forward networks allow signals to travel one way only, from input to output. There is no feedback (loops); that is, the output of any layer does not affect that same layer. Feed-forward neural nets tend to be straightforward networks that associate inputs with outputs \cite{4769258}.

\subsubsection{Convolutional Neural Network (CNN)}
CNN is also a popular and widely used algorithm in deep learning [89]. It has been extensively applied in different applications such as NLP, speech processing, and computer vision [86], for example. Similar to the traditional neural networks, its structure is inspired by the neurons in animal and human brains. Specifically, it simulates the visual cortex in a cat's brain, containing a complex sequence of cells. CNN has three main advantages, namely, parameter sharing, sparse interactions, and equivalent representations. To fully utilize the two-dimensional structure of input data, local connections, and shared weights are employed in the network rather than traditional fully connected networks. This process results in significantly fewer parameters, which makes the network faster and easier to train. This operation is similar to that in visual cortex cells. These cells are sensitive to small sections of a scene rather than the whole scene. In other words, the cells operate as local filters over the input, extracting spatially local correlations that exist in the data \cite{10.1145/3234150}.

\subsubsection{Recurrent Neural Network (RNN)}
Another widely used and popular algorithm in deep learning, especially in NLP and speech processing, is RNN [20]. Unlike traditional neural networks, RNN utilizes the sequential information in the network. This property is essential in many applications where the embedded structure in the data sequence conveys useful knowledge. For example, to understand a word in a sentence, it is necessary to know the context. Therefore, an RNN can be viewed as a set of short-term memory units that comprise the input layer, the hidden (state) layer, and the output layer \cite{10.1145/3234150}.

\subsubsection{Long Short-Term Memory (LSTM and BiLSTM)}
RNNs have difficulties in learning long-term dependencies. The LSTM-based models are an extension of RNNs. The LSTM models extend the RNNs' memory to enable them to keep and learn long-term dependencies of inputs. This memory extension can retain information over a longer period, enabling the reading, writing, and deletion of information from their memories. The LSTM memory is called a "gated" cell, where the word "gate" is inspired by its ability to decide whether to preserve or ignore the memory information. An LSTM model captures important features from inputs and retains this information over an extended period. The decision to delete or preserve the information is made based on the weight values assigned to the information during the training process. Hence, an LSTM model learns which information is worth preserving or removing.

In general, an LSTM model consists of three gates: the forget gate, the input gate, and the output gate. The forget gate determines whether to preserve or remove the existing information, the input gate specifies the extent to which the new information will be added to the memory, and the output gate controls whether the existing value in the cell contributes to the output.
The deep bidirectional LSTMs are an extension of the described LSTM models in which two LSTMs are applied to the input data. In the first round, an LSTM is applied on the input sequence (i.e., forward layer). In the second round, the reverse form of the input sequence is fed into the LSTM model (i.e., backward layer). Applying the LSTM twice leads to improved learning of long-term dependencies and, consequently, will improve the accuracy of the model \cite{unknown}.

\subsubsection{Gated Recurrent Unit (GRU)}
GRU is a type of gated RNN used to address the common problems of vanishing and exploding gradients in traditional RNNs when learning long-term dependencies. They were introduced in 1997 and further improved over the next few years. GRU has an input layer composed of multiple neurons. The size of the feature space determines the number of neurons. Similarly, The output space corresponds to the number of neurons in the output layer. The hidden layer(s) containing memory cells cover the main functions of the GRU networks. Changes and maintenance of cell status depend on two gates in the cell: a reset gate and an update gate. The update gate controls information that flows into memory, and the reset gate controls the information that flows out of memory \cite{SHEN2018895}.

\subsubsection{Attention-based Models}
Attention-based models belong to a model class usually referred to as sequence-to-sequence models. As the name suggests, the purpose of these models is to generate an output sequence based on an input sequence that is usually of different lengths. Based on their performance in computer vision and speech recognition, Attention-based systems were applied to NLP. For tasks such as machine translation \cite{bahdanau2016neural} and text re-construction \cite{rush-etal-2015-neural}, they mainly depend on RNNs and end-to-end encoder-decoders \cite{Yinarticle}. 

\subsubsection{Transformers}
The transformer model introduces an architecture that is solely based on the attention mechanism and does not utilize any recurrent networks. However, it produces results that are superior to those of recurrent networks. It addresses the long-term dependency problem. 
The transformer architecture is also parallelizable, and the training process is considerably faster. Many competitive models have the form of an encoder-decoder. Using the stacked self-attention and point-wise, linear layers for both the encoder and decoder, the Transformer follows this architecture. The encoder maps the input sequence of symbol representations to a continuous sequence of representations. The decoder then produces a sequence of symbol outputs one element at a time. The model is auto-regressive at each step, consuming the symbols previously generated as additional input when generating the next one \cite{46201}.

\subsubsection{Transformer-XL}
Transformer-XL is a novel neural network architecture that enhances the original Transformer design by enabling the model to learn dependencies that span much longer contexts. Compared to the standard Transformer, Transformer-XL offers significantly improved performance in capturing long-term relationships within the data. Additionally, it facilitates more efficient parallelization during computation and eliminates the need for repeated processing of earlier segments, resulting in significantly faster evaluation. By introducing a segment-level recurrence mechanism along with a refined positional encoding scheme, Transformer-XL effectively addresses the issue of fragmented context that often limits the performance of traditional Transformer models \cite{dai-etal-2019-transformer}.

\subsection{Pretrained NLP Models}
\label{sec:pretrained-nlp-models}
In recent years, the field of natural language processing (NLP) has witnessed a shift from task-specific training toward a paradigm based on pretraining followed by fine-tuning. Previously, language models had to be trained from scratch for each downstream task, which was computationally expensive and data-intensive. The development of large-scale, general-purpose pretrained models has enabled significant performance improvements while reducing the need for task-specific resources.

\subsubsection{ULMFiT}
One of the first models to demonstrate the effectiveness of transfer learning in NLP was ULMFiT (Universal Language Model Fine-Tuning for Text Classification) \cite{howard-ruder-2018-universal}. Inspired by techniques used in computer vision, ULMFiT introduced a three-stage pipeline: (1) general-domain language model pretraining, (2) fine-tuning the target task with gradual unfreezing and slanted triangular learning rates, and (3) classification using additional fully connected layers. This approach enabled high performance even with limited labeled data.

\subsubsection{BERT}
BERT (Bidirectional Encoder Representations from Transformers) \cite{devlin2019bert} extended the transfer learning paradigm by introducing deep bidirectional pretraining based on the Transformer architecture. Unlike earlier models that used unidirectional context, BERT learns representations that incorporate both left and right contexts simultaneously. It is trained using two objectives: Masked Language Modeling (MLM) and Next Sentence Prediction (NSP). The same architecture is used during pretraining and fine-tuning, enabling easy adaptation to various downstream tasks such as question answering, sentiment analysis, and natural language inference.

BERT has been released in two primary configurations, BASE and LARGE, differing in depth and model capacity. It has set new benchmarks across multiple NLP tasks and is widely used in both academic research and industry.

\subsubsection{XLNet}
XLNet \cite{XLNet} was proposed to address some of the limitations of BERT, particularly the independence assumptions made during masked language modeling and the discrepancy between pretraining and fine-tuning objectives. XLNet introduces Permutation Language Modeling (PLM), which allows the model to learn bidirectional dependencies without corrupting the input sequence. It also incorporates architectural enhancements from Transformer-XL, such as segment-level recurrence and relative positional encoding, enabling better modeling of long-range context.

Empirical results demonstrate that XLNet outperforms BERT on various benchmark tasks, rendering it a strong alternative for general-purpose language understanding.

\subsection{Evaluation of Text Classifiers}
\label{sec:evaluation}
The efficiency of classification can be evaluated by computing the number of correctly classified class examples (true positives), the number of correctly classified examples that do not belong to the class (true negatives), and examples that either were incorrectly assigned to the class (false positives) or that were not recognized as class examples (false negatives). These four counts constitute a confusion matrix, shown in Table \ref{tab:Confusion matrix}, for the case of binary classification \cite{measures}.

\begin{table}[t!]
	\centering
	\caption{Confusion matrix for binary classification}
	\label{tab:Confusion matrix}
	\begin{tabular}{lll}
		\toprule
		\textbf{Data class} & \textbf{Classified as Pos} & \textbf{Classified as Neg}   \\
		\midrule
		\textbf{Pos} & True Positive (TP) & False Negative (FN)  \\
		\textbf{Neg} & False Positive (FP) & True Negative (TN)  \\  
		\bottomrule
	\end{tabular}
\end{table}

The most often used measures for binary classification based on the values of the confusion matrix are:

\begin{itemize}
	\item{Accuracy:} the percentage of texts that have been projected with the correct tag is a classifier's overall effectiveness and can be determined as follows:
	\begin{align*}
		Accuracy &= \frac{TP + TN}{TP + FN + FP + TN}\\
	\end{align*}
	
	\item{Precision:} the percentage of samples the classifier predicted correctly out of the total number of examples that it predicted for a given tag, which is calculated as below:
	\begin{align*}
		Precision &= \frac{TP}{TP+FP}\\
	\end{align*}
	\item{Recall:} the percentage of examples that the classifier predicted for a given tag out of the total number of examples for that given tag that it should have predicted:
	\begin{align*}
		Recall &= \frac{TP}{TP+FN}\\
	\end{align*}
	\item{F1 Score:} Taking the weighted harmonic mean of precision and recall leads to F1-Score\cite{monkeylearn}\cite{measures}:
	
	\begin{align*}
		F1 = \frac{2 \cdot precision\cdot recall}{precision+ recall}\\
	\end{align*}
\end{itemize}

These evaluation metrics provide essential insight into the performance of text classification models. They help quantify different aspects of predictive quality and guide model selection and tuning. A well-rounded assessment typically involves striking a balance between precision and recall, depending on the specific requirements of the application domain.

\subsection{Summary}
Various machine learning methods that have been used for text classification are discussed in this section. Deep learning models have shown to be more reliable comparing to classical approaches for text classification, thus they have attracted the attention of the researchers in recent years. The next section presents in details the most important approaches for text classification.

\section{Related work}
\label{sec:related-work}
The rapid growth of computing and networking has led to new applications in text classification, resulting in increased demands for its effectiveness. However, text classification presents several challenges for learning systems. The feature vector or structure used to represent text must adequately capture the complex semantics of natural language. Additionally, the features should be applicable across a wide range of class definitions, as new features typically cannot be developed for each user or category.

Hotho et al. \cite{surveyoftextmining} present four major challenges in classifying text generated by social media: short texts, abundant information, noisy phrases, and time sensitivity. Short texts contain very few words, resulting in a lack of features for classification, which can negatively impact the results. For instance, Twitter generates around 500 million tweets daily in various languages, including both spam and personal conversations. This necessitates filtering tweets for spam and noise prior to analysis. However, the high volume of tweets means that pre-processing and classification require significant time, making real-time classification challenging \cite{CaseStudy}.

Many recent studies have been conducted in the fields of natural language processing and machine learning to address these issues. This section discusses recent advancements in these areas.

\subsection{Instance-based Algorithms}
The instance-based learning model is designed as a decision-making framework that relies on training data examples. These methods typically compile a database of example data and use a similarity measure to compare new data against this database, aiming to find the best match for making predictions \cite{machinelearningmastery}.

Alakrot et al. \cite{TowardsAccurate} propose a predictive model for detecting anti-social behavior on online platforms in Arabic. They gathered and labeled a large dataset from YouTube comments, which included numerous offensive and non-offensive comments. During the pre-processing phase, the researchers first separated the alphabetic sequences into tokens. Next, they removed commas, punctuation marks, diacritics in Arabic, and select words from the documents. This was followed by normalization, where certain letters were replaced with others to enhance text mining operations. 

For classification, an instance-based classifier using Support Vector Machine (SVM) was built with word-level features. To evaluate the classifier's accuracy, the dataset was divided into training and test sets, and 10-fold cross-validation was applied. The algorithm performed well after implementing data pre-processing with stemming, achieving an F1-score of 0.82.

\subsection{Bayesian Algorithms}
Bayesian methods are those that explicitly apply Bayes’ Theorem for classification problems \cite{machinelearningmastery}. Shende et al. \cite{Acomputationalframework} propose an offensive language classification system for social communities using supervised learning techniques. This system can identify potential users responsible for spreading offensive language in discussions on platforms like YouTube and Facebook. 

In the preprocessing stage for unstructured data, stop words are removed, and slang is mapped to its original form. Following this, various data mining techniques are employed for feature extraction, including tokenization, term frequency, inverse term frequency, and the N-gram technique. The system consists of two phases: first, it detects offensive content, and second, it identifies potential users.

Support Vector Machine (SVM) and Naive Bayes (NB) classifiers are utilized across the overall discussion. LIBSVM, the library used for SVM in Weka, does not understand sentence structures, so the classification of each post is instead numerical: 1 indicates a non-offensive post, while -1 indicates an offensive post. Potential users are identified based on the following equation for each user:

\begin{equation}
U^i = \frac{\text{Number of detected offensive posts by classifier}}{\text{Total number of posts made by user}}
\end{equation}

If the calculated value exceeds a certain threshold, the user is labeled as a potential source of offensive language. The experimental results for offensive post classification demonstrate high accuracy for both classifiers, with SVM achieving 91.75\% accuracy and the Naive Bayes algorithm achieving 90\%.

\subsection{Decision Tree Algorithms}
Decision tree methods create models based on actual values of attributes in the data. These decision paths expand in tree structures until a prediction is made for a given record \cite{machinelearningmastery}.

Muhammad Okky et al. \cite{Indonesian} discuss offensive language detection in the Indonesian language on social media. They developed a new Twitter dataset specifically for detecting abusive language in Indonesian tweets. The first step in their experiments involved crawling Twitter data using the Twitter API and the Tweepy Library. Afterward, they filtered the data by removing duplicate tweets and those written in local or foreign languages. The dataset was then labeled into three categories: abusive but not offensive, not abusive, and offensive language.

Next, they optimized the text by removing unimportant features such as usernames, retweets, and URLs. They also eliminated hashtags, punctuation, and emoticons from the tweets. Following this, they mapped nonformal words to their formal equivalents using a dictionary. The next step involved feature extraction, where they utilized word n-gram features and character n-grams. They employed several classifiers, including Naive Bayes (NB), Support Vector Machines (SVM), and Random Forest Decision Trees (RFDT), implemented using Scikit-Learn in Python. They applied a 10-fold cross-validation technique for evaluation, using 90\% of the data for training and 10\% for testing. The F1 score was used as the evaluation metric.

Two experimental scenarios were designed: the first classified tweets into the three labels mentioned earlier, while the second classified the dataset into two labels: abusive and non-abusive language. For word n-grams, they utilized unigrams, bigrams, trigrams, and combinations of all three, while for character n-grams, they used trigrams, quadgrams, and combinations of character quadgrams and trigrams. The results indicated that classifying tweets into three labels was more challenging than classifying them into two classes, primarily due to the difficulty in distinguishing between tweets that are abusive but not offensive and those that are offensive. Overall, the experimental results demonstrated that Naive Bayes outperformed both SVM and RFDT.

Jorge et al. \cite{Anomaly} propose an anomaly-detection method for identifying trolling comments on social news websites. The filtering method is based on several features. They used the 'Meneame' dataset, a Spanish social news website where news and stories are published. The extracted features from the comments were categorized into three groups: syntactic, statistical, and opinion-based. They employed an anomaly detection approach using these features to represent comments as points in a feature space. This method allowed them to identify a group of comments that represented normal behavior (troll comments) and to determine whether a particular comment is categorized as Not Troll or Troll by measuring its deviation from the reference group.

To assess the similarity between different comments, they computed Euclidean, Manhattan, and Cosine distances. They conducted a 5-fold cross-validation and applied various supervised machine learning algorithms, including Bayesian networks, SVM, K-nearest neighbors, and decision trees. When comparing anomaly classification of comments on social news websites with the results from supervised machine learning approaches, they achieved comparable results. The best-performing supervised machine learning algorithm obtained an F-measure of 90.72\%, whereas anomaly detection achieved 76.13\%.

\subsection{Deep Learning Algorithms}
Deep learning methods represent a modern evolution of artificial neural networks, leveraging the availability of abundant and affordable computation resources \cite{machinelearningmastery}.

Abubakar et al. \cite{PROFILING} propose a method based on a user profiling algorithm that employs a deep Long Short-Term Memory (LSTM) neural network trained to detect abusive language. Two distinct datasets were created: one for abusive language detection and another for user profiling. The user profiling dataset comprised tweets from users who posted abusive messages, along with their account names. These tweets were labeled according to corresponding abusive language categories, including bigots, racists, extremists, and offensive individuals. The abusive language detection dataset was compiled using tweets across all abusive language categories obtained via the Twitter API. Subsequently, the dataset was cleaned and transformed into a format suitable for learning algorithms. This cleaning process involved removing noise and unwanted features such as URLs, stop words, emoticons, and special characters (including any characters not in the a-z range). Afterward, the dataset was tokenized. During the feature extraction and representation phase, word embeddings were selected as the feature representation method.

The user profiling algorithm analyzes the frequency of the labels assigned to each user's tweets and their connections as detected by the abusive language detection model. It determines the label with the highest number of occurrences for each user (user polarity) and evaluates the connections of users (both followers and followees) to establish the mode of these connections (following polarity and follower polarity). The algorithm relies on the LSTM recurrent neural network model to classify the tweets. The LSTM layer captures long-distance semantic and contextual information effectively. The dataset was divided into three parts for the classification experiment: 80\% for training, 10\% for validation, and 10\% for testing. The LSTM-based neural network, trained with both pretrained word embeddings and word embeddings learned through the embedding layer, achieved validation accuracies of 88.84\% and 88.62\%, respectively.

Rosa et al. \cite{Deeper} implement three architectures for cyberbullying detection: a simple Convolutional Neural Network (CNN), a hybrid CNN-LSTM model, and a mixed CNN-LSTM-Deep Neural Network (DNN). They utilize an extended version of the Formspring dataset available on Kaggle, which is based on a question-and-answer format. As a preprocessing step, they remove the “Q:” and “A:” markers, any encoding representations, and HTML tags. Their experimental setup consists of three steps: First, they implement the CNN as detailed by Kim et al. \cite{Convolutional}, which comprises a one-layer CNN followed by a fully connected layer with a dropout of 0.5 and a softmax output. Next, they implement the hybrid CNN-LSTM model as described by Zhou et al. \cite{CLSTM}, featuring a CNN layer followed by an LSTM layer. Lastly, they apply the mixed CNN-LSTM-DNN model from Ghosh and Veale \cite{Fracking}. Cross-validation (10-fold) is conducted, and parameters are optimized for each fold. They test three distinct text representations: Google News word embeddings, Formspring word embeddings, and Twitter word embeddings. Their results indicate that, in the unbalanced dataset, the CNN model by Kim et al. using Google embeddings achieved the best performance with an F-measure of 84.8\%. In contrast, for the balanced dataset, the C-LSTM model by Zhou et al. using Twitter embeddings achieved the best result, with an F-measure of 84.2\%.

Abubakar et al. \cite{COMPARATIVE} evaluate the performance of traditional machine learning algorithms and deep learning algorithms on the task of tweet classification. The pre-processing steps include removing URLs, emoticons, special characters, and stop words from the dataset, followed by tokenization. The output variable is categorical and indicates the class to which each tweet belongs. After pre-processing, labels were manually assigned to each tweet. There are five classes: 0 denotes bigotry, 1 denotes offensive language, 2 denotes racism, 3 indicates extremist views, and 4 indicates that a tweet does not contain any abusive language. After data pre-processing, features are extracted for classification by the learning algorithms. 

The two types of feature representations used in the experiment are Bag of Words (BoW) and Word Embeddings. In BoW, keywords are filtered from the training data using various NLP methods. Word embeddings provide learning algorithms with syntactic and semantic information by grouping similar words together in a vector space. The researchers applied several machine learning algorithms, including Naïve Bayes, Support Vector Machines (SVM), Random Forest, K-Nearest Neighbors (KNN), and Logistic Regression, as well as deep learning algorithms like Long Short-Term Memory (LSTM) and Gated Recurrent Units (GRU). GRU is an updated form of LSTM that maintains LSTM's resistance to the vanishing gradient problem. 

The results indicate that four of the models performed well in classifying tweets into the five categories, except in the cases of bigotry and racism. Naïve Bayes achieved an accuracy of 62\%, SVM achieved 71\%, Logistic Regression reached 70\%, and GRU achieved 88\%, outperforming LSTM which obtained an accuracy of 87.16\%. KNN was the worst-performing model, recording the lowest accuracy rates of 47\% and 56\%.

A paper categorizing offensive language in social media, presented in 2019, is based on results from an international workshop in semantic evaluation called SemEval, which took place in 2019. This competitive workshop included 12 different tasks to be solved, with the current work focusing on the results of Task No. 6, which involved identifying and categorizing offensive language in social media. This task featured three sub-tasks: identifying offensive language, categorizing the types of offenses, and identifying the targets of the offenses. It utilized a new dataset called the Offensive Language Identification Dataset (OLID), composed of 14,000 English tweets. In total, 800 teams participated in this task, with 115 successfully submitting their results. The outcomes were compared using a macro-averaged F1 score on the test subset of the dataset.

Approximately one third of the teams performed some form of preprocessing, which included standardizing tokens, hashtags, URLs, retweets, dates, elongated phrases, and even partially concealed words. Other preprocessing strategies involved converting emojis to text, deleting uncommon words, and using Twitter-specific tokenizers.

While various types of systems were presented, Figure 3.1 indicates that the most popular model type for sub-task A was Ensemble methods. However, the evaluation results show that the best-performing systems used ensembles and state-of-the-art deep learning models like BERT. In fact, the top-performing team, along with seven out of the top ten teams, employed BERT. The highest F1 score achieved on the test set was 82.9\% \cite{Liu2019NULIAS}.

In sub-task B, which involved offense type categorization, ensembles also emerged as the top-performing method, with 50\% of the top ten teams utilizing some form of ensemble technique. The leading team in this sub-task employed a rule-based approach, achieving a 75.5\% F1 score, while finishing in 76th place in the previous sub-task \cite{zampieri2019semeval}.

For the target identification sub-task, half of the top ten teams again used ensemble methods, with the winning team utilizing the pretrained BERT model \cite{nikolov-radivchev-2019-nikolov}.

Although the BERT model secured two first-place finishes in the sub-tasks, the best team overall (with an average rank of 6.33) ranked 2nd, 16th, and 1st across the sub-tasks, indicating that no single model achieved out-performing results across all tasks simultaneously.

\subsection{Ensemble Algorithms}
Ensemble methods are models composed of multiple weaker models that are independently trained, and their predictions are combined to make an overall decision \cite{machinelearningmastery}.

Chin et al. \cite{Lyrics} propose an automated approach for screening song lyrics to detect harmful content. They used the South Korean Broadcasting System dataset to classify songs as "Unqualified Fail" for broadcasting if their lyrics included explicit terms related to sex, drugs, or profanity, and "Pass" for broadcast if the lyrics did not contain such terms. The dataset includes song titles, musician names, and the broadcast outcome (Pass or Fail). The baseline of the experiment involved a method that classified songs as "Fail" or "Pass" by detecting explicit content using a profanity language dictionary released by Namu-wiki. They applied two well-known ensemble algorithms, Bagging and AdaBoost. The lyrics were transformed into a vector, with the dimension determined by the vocabulary size of the training data. They compared three forms of the vector: the first used the entire vocabulary with term frequency as the vector value; the second used the whole vocabulary with the Inverse Document Frequency (IDF) score, ignoring term frequency; and the third used a selective vocabulary based on 'parts of speech' tagging with the IDF score as the vector value. The results showed that the effectiveness of each model increased when the IDF score was used. The best-performing model, in terms of all performance metrics, was the Bagging model that employed selective terms based on POS tagging along with the IDF vector.

Juan et al. \cite{vandalism} implemented a vandalism detection system for Wikipedia that utilizes three sets of features based on different approaches. Vandalism is defined as any deletion, addition, or alteration of content made intentionally to compromise the integrity of Wikipedia. The first set of features generalizes vandalism term lists, leveraging the semantic similarity relationships found in dense word representations, known as word embeddings. This results in four features, two focusing on frequency and impact, and the other two on words found in a vandalism cluster. The second set of features is derived from a deep neural network (SDA network), producing 25 features based on different combinations of neurons in the three input layers. Finally, the third set generates eight features using graph-based ranking algorithms (co-occurrence graph). Four types of measures were applied to rank the generated vandal graph: degree centrality, eigenvector centrality, and PageRank. In their experiments, a wide range of classification algorithms were analyzed, but the Random Forest algorithm performed the best, achieving an F-measure of 79.1\%.

\subsection{Discussion}
The literature review shows that BERT has been widely used for offensive language detection tasks, particularly those based on the OLID dataset. Numerous submissions to the SemEval 2019 Task 6, including those by Liu et al. \cite{Liu2019NULIAS}, highlighted BERT’s strong performance in subtasks A and B. Indeed, most of the top-ranking systems in the competition utilized BERT or its variants as the primary model for classification.

In contrast, despite its theoretical advantages, XLNet has not yet been applied to the OLID dataset or its hierarchical subtasks (A, B, and C) in existing studies. Although XLNet has been investigated in broader contexts of offensive or abusive language detection, there has been no systematic evaluation of its performance on the OLID benchmark. This leaves a significant gap in the research: while XLNet’s permutation-based autoregressive architecture offers improvements over BERT in capturing bidirectional context and long-range dependencies, its potential for hierarchical offensive language classification has yet to be explored.

Our contribution aims to address this gap by being the first to apply XLNet to the OLID subtasks and compare its performance against BERT, which serves as a well-established baseline. Additionally, we examine the impact of oversampling on both models, paying particular attention to underrepresented classes, such as 'OTH' in subtask C. Through this approach, we seek to determine whether XLNet’s theoretical advantages translate into practical improvements in this challenging classification task.

\section{Design}
\label{sec:design}
Social media platforms allow users to express their thoughts without text constraints. As a result, this freedom can lead to the use of offensive language in their posts. To tackle this issue, text classification is essential. Our literature review indicates that deep learning models have demonstrated reliability in text classification tasks. Therefore, in this work, we propose a method for detecting offensive language using the XLNet model \cite{XLNet}.

This section begins with a summary of the proposed method, followed by a detailed explanation of the approach. The final section outlines the experimental setup.

\subsection{Overview}

Detecting offensive language on Twitter involves a three-level hierarchical classification schema based on the work of Zampieri et al. \cite{zampierietal2019}. The categories in this schema are: offensive language detection, categorization of offensive language, and identification of the target of the offensive language. We propose applying this framework to the Offensive Language Identification Dataset (OLID)\footnote{https://sites.google.com/site/offensevalsharedtask/offenseval2019}, a publicly available dataset of English tweets.

Each category addresses a specific type of offensive content but shares common characteristics that the hierarchical annotation model aims to capture. For example, insults directed at individuals are typically classified as cyberbullying, while insults aimed at groups fall under hate speech. The hierarchical annotation scheme utilized by OLID makes it a valuable resource for various tasks related to recognizing and characterizing offensive language.

For our work, we propose using XLNet \cite{XLNet}, which has demonstrated superior performance over BERT \cite{devlin2018bert} across twenty NLP tasks, including question answering, natural language inference, and sentiment analysis. To the best of our knowledge, XLNet has not yet been applied to the task of offensive language detection.

\subsection{Dataset}
Data selection is the first step of any detection system. As mentioned earlier, we intend to work on the OLID dataset, which contains more than 14,000 tweets. It has been annotated using a hierarchical three-level annotation model published in 2019 \cite{zampierietal2019}.

For anonymization and normalization purposes, user-specific data was removed. The dataset was sampled such that 50\% of the tweets came from political keywords and the other 50\% from non-political keywords, as political content tends to be richer in offensive language.

Table~\ref{tab:Dataset} demonstrates the breakdown of the data into training and testing sets for each category label. The OLID dataset was chosen due to its hierarchical structure and its use in major NLP events, such as SemEval-2019 \footnote{The dataset was created explicitly for SemEval-2019 Task 6: Detecting the type and target of offensive content.}.

\begin{table}[t!]
	\centering
	\caption{Distribution of label combinations in OLID}
	\label{tab:Dataset}
	\begin{tabular}{lllrrr}
		\toprule
		\textbf{A} & \textbf{B} & \textbf{C} & \textbf{Training} & \textbf{Testing} & \textbf{Total} \\ \midrule
		OFF        & TIN        & IND        &             2,407 &              100 &          2,507 \\
		OFF        & TIN        & OTH        &               395 &               35 &            430 \\
		OFF        & TIN        & GRP        &             1,074 &               78 &          1,152 \\
		OFF        & UNT        & —          &               524 &               27 &            551 \\
		NOT        & —          & —          &             8,840 &              620 &          9,460 \\ \midrule
		All        &            &            &            13,240 &              860 &         14,100 \\ \bottomrule
	\end{tabular}
\end{table}

\subsubsection{Hierarchical Annotation of Offensive Content}
\begin{figure*}[t!]
	\centering
	\includegraphics[width=0.9\linewidth]{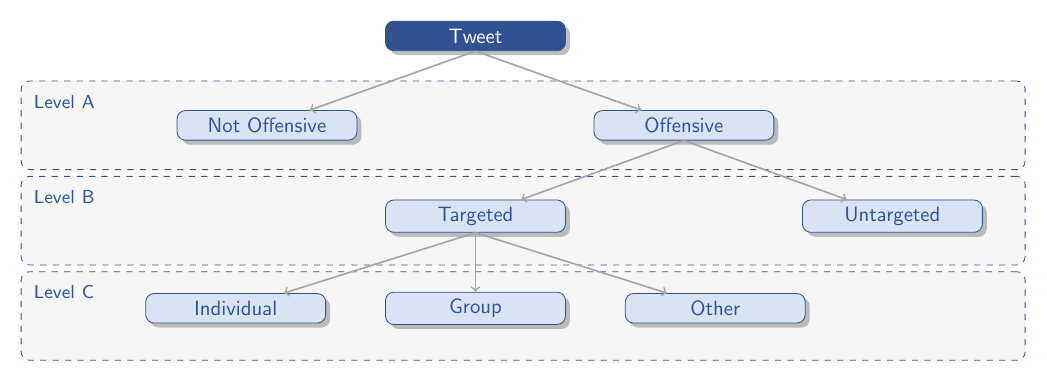}
	\caption{Hierarchical annotation schema used in the OLID dataset.}
	\label{fig:classification-levels}
\end{figure*}

The OLID dataset is annotated at three levels: whether the content is offensive (Level A), the type of offense (Level B), and the target of the offense (Level C), as shown in Figure~\ref{fig:classification-levels}. Table~\ref{tab:OLID Sample} provides examples illustrating this classification scheme.

\begin{table*}[ht!]
	\centering
	\caption{Sample from the OLID dataset, including tweet text and hierarchical labels}
	\label{tab:OLID Sample}
	\begin{tabular}{p{9cm}lll}
		\toprule
		\textbf{Tweet}                                                      & \textbf{A} & \textbf{B} & \textbf{C} \\ \midrule
		@USER He is so generous with his offers.                            & NOT        & -          & -          \\
		IM FREEEEE!!!! WORST EXPERIENCE OF MY F***ING LIFE                  & OFF        & UNT        & -          \\
		@USER F** this fat **** sucker                                      & OFF        & TIN        & IND        \\
		@USER Figures! What is wrong with these idiots? Thank God for @USER & OFF        & TIN        & GRP        \\ \bottomrule
	\end{tabular}
\end{table*}

\begin{itemize}
	\item \textbf{Level A (Offensive Language Identification):} classifies tweets into the following types:
	\begin{itemize}
		\item \textbf{Not Offensive (labelled as NOT):}	Tweets that do not contain any offensive content.
		
		\item \textbf{Offensive (labelled as OFF):} Tweets containing targeted offense or unacceptable language, including insults, threats, or profanity.
	\end{itemize}
	
	\item \textbf{Level B (Offensive Language Categorization):} categorizes the type of offense:
	\begin{itemize}
		\item \textbf{Targeted Insult (labelled as TIN):} Tweets containing insults or threats directed at an individual or group.
		
		\item \textbf{Untargeted (labelled as UNT):} Tweets containing general profanity not directed at a specific individual or group.
	\end{itemize}
	
	\item \textbf{Level C (Offensive Language Target Identification):} identifies the target of the offense in targeted tweets:
	\begin{itemize}
		\item \textbf{Individual (labelled as IND):} Tweets targeting a specific individual, named or unnamed, often associated with cyberbullying.
		
		\item \textbf{Group (labelled as GRP):} Tweets targeting a group based on shared characteristics such as race, gender, religion, etc., often considered hate speech.
		
		\item \textbf{Other (labelled as OTH):} Tweets targeting entities not classified as individuals or groups, such as organizations, events, or abstract concepts.
	\end{itemize}
\end{itemize}

\subsection{Classification Pipeline}

Our proposed system is divided into three phases: the preprocessing phase, the training phase, and the evaluation phase.

\subsubsection{Preprocessing Phase}
Preprocessing tweets is essential to remove irrelevant content and reduce dataset complexity. Tweets are first cleaned by removing mentions (@user), URLs, and most punctuation marks, as these elements do not contribute useful information for classification. The symbols `@` and `\#` are preserved due to their semantic significance in tweets.

Next, the tweets undergo tokenization and lowercasing. Tokenization is the process of splitting a string into smaller units (tokens) to enhance text understanding. For example, the phrase \textit{Offensive Language Detection} is split into the tokens \textit{Offensive}, \textit{Language}, and \textit{Detection}; applying lowercasing yields \textit{offensive}, \textit{language}, and \textit{detection}.

Hashtag tokens are split into words if they are either camel-cased (e.g., \#StopHate) or separated by underscores (e.g., \#stop\_hate). Emojis are replaced with their textual descriptions using Python's \texttt{emoji} library, and the original emoji symbols are removed.

\subsubsection{Tweet Classification}

After preprocessing, rich textual representations are obtained using pre-trained language models, which have significantly advanced performance across a wide range of natural language processing (NLP) tasks \cite{embeddings}.

\begin{itemize}
	
	\item \textbf{Transformer Models.}  Pre-trained models based on the Transformer architecture have become foundational in modern NLP. Transformers enable large-scale pretraining on unlabeled corpora followed by fine-tuning on specific tasks. Unlike traditional sequential models such as LSTMs, Transformers process entire input sequences in parallel using self-attention mechanisms, resulting in improved efficiency and accuracy. A generic Transformer model includes the following components:	
	\begin{enumerate}
		\item \textbf{Inputs:} Raw text input.
		\item \textbf{Input Embedding:} Tokens are mapped to dense vectors using embedding models such as Word2Vec or GloVe.
		\item \textbf{Positional Encoding:} Positional information is added to embeddings to preserve word order.
		\item \textbf{Multi-Head Attention:} Self-attention is applied in parallel across multiple heads to capture diverse dependencies.
		\item \textbf{Add and Norm:} Residual connections and layer normalization are applied.
		\item \textbf{Feed-Forward Layer:} Dense layers further transform the representations.
		\item \textbf{Decoder and Output:} Final layers produce output vectors, which are projected and normalized (e.g., via Softmax).
	\end{enumerate}
	
	\item \textbf{BERT} uses only the encoder portion of the Transformer architecture. It is pre-trained on large text corpora using two unsupervised objectives:
	\begin{itemize}
		\item \textbf{Masked Language Modeling (MLM):} Randomly masks 15\% of the input tokens and predicts them using context.
		\item \textbf{Next Sentence Prediction (NSP):} Predicts whether a given sentence logically follows another.
	\end{itemize}
	
	Two model configurations are commonly used: BERT-BASE (12 layers, 768 hidden units, 12 attention heads) and BERT-LARGE (24 layers, 1024 hidden units, 16 attention heads).
	
	In our classification architecture, the input tweet is tokenized and passed to BERT. The representation of the special \texttt{[CLS]} token, which captures global sentence-level semantics, is then fed into a feed-forward neural network followed by a Softmax layer. This pipeline is illustrated in Figure~\ref{fig:bert-classification}.
	
	\begin{figure*}[ht!]
		\centering
		\includegraphics[width=\linewidth]{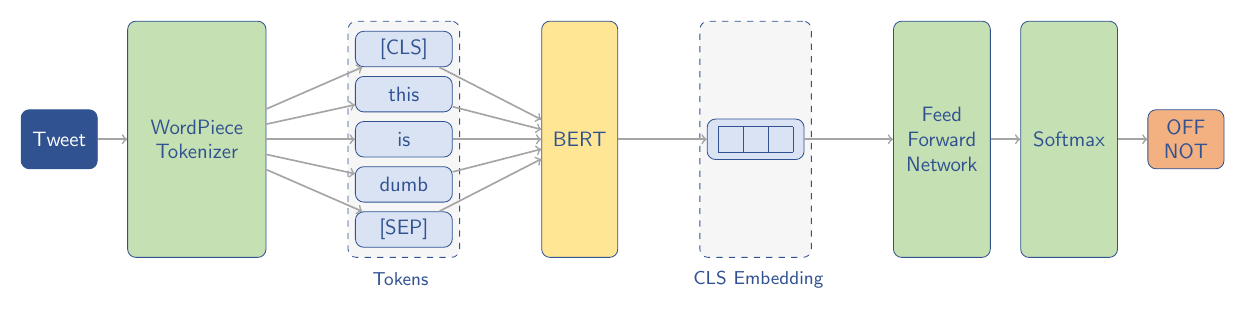}
		\caption{BERT-based architecture}
		\label{fig:bert-classification}
	\end{figure*}
	
	\item \textbf{XLNet} \cite{XLNet} is an autoregressive language model that addresses several limitations of BERT. Instead of masking inputs, XLNet employs Permutation Language Modeling (PLM), training on all possible permutations of token order. This allows the model to capture bidirectional context without corrupting the input signal.
	
	XLNet also incorporates mechanisms from Transformer-XL, including segment-level recurrence and relative positional encodings, which allow it to model longer dependencies effectively.
	
	Available model sizes include:
	\begin{itemize}
		\item \textbf{XLNet-BASE:} 12 layers, 768 hidden units, 12 attention heads, 110M parameters.
		\item \textbf{XLNet-LARGE:} 24 layers, 1024 hidden units, 16 attention heads, 340M parameters.
	\end{itemize}
	
	In our system, the output vectors corresponding to all tokens are aggregated using mean pooling to obtain a fixed-length representation of the input. This is then passed through a fully connected classification layer. The architecture is shown in Figure~\ref{fig:xlnet}.
	
	\begin{figure*}[ht!]
		\centering
		\includegraphics[width=\linewidth]{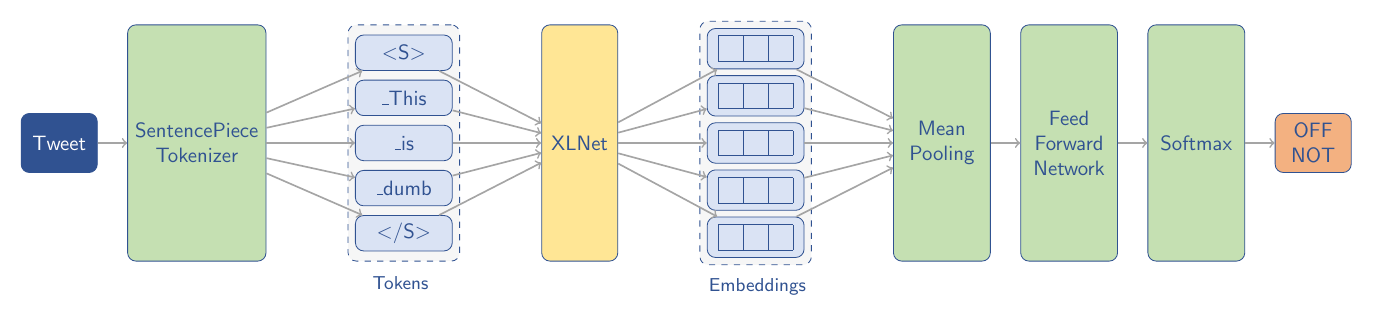}
		\caption{XLNet-based architecture}
		\label{fig:xlnet}
	\end{figure*}
	
	\item \textbf{Comparison Between XLNet and BERT.}
	
	Consider the sentence ''New York is a city.'' If both models aim to predict the phrase ``New York,'' their computations differ due to the underlying training objectives:
	\begin{multline}
		\text{BERT} = \log P(\text{New} \mid \text{is a city}) \\+ \log P(\text{York} \mid \text{is a city})
	\end{multline}
	\begin{multline}
		\text{XLNet} = \log P(\text{New} \mid \text{is a city}) \\+ \log P(\text{York} \mid \text{New, is a city})
	\end{multline}
	
	XLNet captures dependencies such as the bigram (``New'', ``York'') by conditioning on all permutations, while BERT masks tokens independently and does not model their order during training. Although BERT may still learn useful patterns like (``New'', ``city'') and (``York'', ``city''), XLNet yields denser training signals and models contextual dependencies more effectively.
	
	In this work, we propose an automatic system for offensive language detection in tweets based on XLNet and compare its performance to that of BERT. To the best of our knowledge, XLNet has not been previously applied to this specific task.
	
\end{itemize}

\section{Experimental Results}
\label{sec:experimental}
This section presents the experiments conducted to evaluate the performance of the proposed approach. It includes a description of the dataset, preprocessing techniques, experimental setup, model training, and evaluation metrics.

\subsection{Dataset}
As described in Section~\ref{sec:design}, the Offensive Language Identification Dataset (OLID)\footnote{\url{https://sites.google.com/site/offensevalsharedtask/offenseval2019}}, which contains 14,100 English tweets, was used in the experiments. The dataset, introduced by Zampieri et al.~\cite{zampierietal2019}, is annotated using a three-level hierarchical schema: (A) whether the tweet is offensive, (B) whether the offense is targeted, and (C) the target of the offense (individual, group, or other).

The dataset is split into 80\% for training and 20\% for testing. Since OLID is imbalanced across classes, two resampling techniques were applied independently to balance the training data: 
\begin{itemize}
	\item \textbf{Undersampling:} Randomly removing examples from the majority class to match the size of the minority class.
	\item \textbf{Oversampling:} Randomly duplicating examples from the minority class (with replacement) to balance the dataset.
\end{itemize}

\subsection{Tweet Preprocessing}
Preprocessing helps reduce noise and improve model performance while also decreasing training time. The following techniques were applied to the tweets in both training and testing sets:
\begin{itemize}
	\item \textbf{Tweet Cleaning:} Mentions (@user) and URLs were removed, as they do not contribute meaningful content.
	
	\item \textbf{Lowercasing:} All tweets were converted to lowercase to normalize the text.
	
	\item \textbf{Emoji Substitution:} Emojis were replaced with their corresponding textual descriptions using the \texttt{emoji} Python library, and the original emoji symbols were removed.
	
	\item \textbf{Punctuation Removal:} Most punctuation marks were removed since they do not enhance the semantic understanding of tweets.
	
	\item \textbf{Hashtag Segmentation:} Hashtags written in camel case (e.g., \#StopHate) or separated by underscores (e.g., \#stop\_hate) were segmented into constituent words (e.g., ``stop hate'').
\end{itemize}

\subsection{Experimental Setup}
The experiment compares the performance of XLNet and BERT on the three classification tasks (A: offensive detection, B: offense categorization, C: target identification).
\begin{itemize}
	\item \textbf{Hardware:}
	\begin{itemize}
		\item Laptop with Intel Core i7 CPU and 16 GB RAM.
		\item Google Colab with Tesla P4 GPU.
	\end{itemize}
	
	\item \textbf{Software:}
	\begin{itemize}
		\item Python programming language.
		\item PyTorch-based implementation of XLNet and BERT using Hugging Face's Transformers library\footnote{\url{https://huggingface.co/transformers}}.
	\end{itemize}
\end{itemize}

\subsection{Model Training}
To ensure a fair comparison, both XLNet and BERT models were trained using the same dataset for each task. We selected the \texttt{Base} versions of both models, which share the same architecture size—12 layers, 768 hidden units, and 12 attention heads. Training configurations such as batch size and number of epochs were kept consistent across models, while minor adjustments were made to learning rates to suit each model's optimization needs. Details for each model’s training setup are provided below.

\subsubsection{Hyperparameters of BERT}
We used BERT-Base for similar reasons. It has the same number of layers, hidden units, and attention heads as XLNet-Base. The AdamW optimizer was also used, with a learning rate of 5e-5 and a weight decay of 0.01. Batch size and number of epochs were kept consistent with the XLNet configuration.

\subsubsection{Hyperparameters of XLNet}
We used XLNet-Base due to GPU memory constraints. This configuration includes 12 layers, 768 hidden units, and 12 attention heads. The model was trained using the AdamW optimizer with a learning rate of 2e-5 and a weight decay of 0.01. The batch size was set to 32. The number of epochs was tuned from 1 to 10, and in all subtasks, the best validation loss was achieved at 3 epochs or fewer. The maximum input length was set to 150 tokens, matching the maximum tweet length in the dataset.

\subsection{Model Evaluation}
\label{sec:evaluation_results}

Both models were evaluated on the same test set using the standard classification metrics: accuracy, precision, recall, and F1 score, which were previously defined in Section~\ref{sec:evaluation}. 

We report results per class as well as averaged across all classes. The primary metric used for comparison is \textbf{Macro-F1}, which calculates the unweighted average of F1 scores across all classes. This choice is motivated by the class imbalance in the dataset, where Macro-F1 better reflects the model’s performance on minority classes \cite{Liu2019NULIAS}.

\subsubsection{Experimental Results}
We present the performance of BERT and XLNet on the test set in Table~\ref{tab:bert_all_subtasks} and Table~\ref{tab:xlnet_all_subtasks}, respectively. It is worth noting that for subtasks A and B, the overall ("ALL") metrics closely mirror the performance of the majority class (e.g., "NOT" in subtask A and "TIN" in subtask B). This is a typical outcome in imbalanced classification tasks, where the large number of examples in the dominant class disproportionately influences aggregate metrics such as accuracy and micro-averaged F1. A

For subtask A, XLNet outperformed BERT, achieving a Macro-F1 of 0.78 compared to 0.69. It also achieved higher accuracy (0.81 vs. 0.80), indicating better overall classification of offensive vs. non-offensive tweets.

In subtask B, XLNet again performed better, with a Macro-F1 score of 0.71 and an accuracy of 0.83, compared to BERT's scores of 0.63 and 0.78, respectively. This suggests that XLNet was more effective in identifying whether offensive language was targeted.

However, in subtask C, both models struggled—especially with the "OTH" (other) class, where neither model predicted any instances correctly. XLNet's overall Macro-F1 for this subtask was 0.36, while BERT slightly outperformed it with a Macro-F1 of 0.37 and a higher accuracy (0.45 vs. 0.56). This suggests that neither model was reliable in identifying the specific target of offensive language when it fell outside common categories, such as individuals or groups.

Oversampling had a notable impact, particularly for subtask B. XLNet showed marked improvement, reaching an accuracy of 0.86 and an F1 score of 0.92. In contrast, subtask A exhibited only minor improvements for both models. For BERT, oversampling led to an accuracy increase to 0.85 in subtask A and consistent performance gains in subtask B.

Overall, subtask A was the easiest to classify, followed by subtask B. Subtask C remained challenging due to the sparsity of examples in the "OTH" class. Across all tasks, XLNet demonstrated stronger performance than BERT on subtasks A and B, while BERT performed marginally better in subtask C. Additionally, oversampling was generally more beneficial than undersampling for both models.

\begin{table*}[ht!]
	\centering
	\caption{BERT results on all subtasks (A, B, and C)}
	\label{tab:bert_all_subtasks}
	\begin{tabular}{lccccc}
		\toprule
		\textbf{Subtask A} & \textbf{Precision} & \textbf{Recall} & \textbf{F1 score} & \textbf{Accuracy} & \textbf{MacroF} \\
		\midrule
		NOT & 0.80 & 0.98 & 0.88 & -- & -- \\
		OFF & 0.87 & 0.35 & 0.50 & -- & -- \\
		ALL & 0.87 & 0.35 & 0.50 & 0.80 & 0.69 \\
		\midrule
		\textbf{Subtask B} & \textbf{Precision} & \textbf{Recall} & \textbf{F1 score} & \textbf{Accuracy} & \textbf{MacroF} \\
		\midrule
		UNT & 0.29 & 0.67 & 0.40 & -- & -- \\
		TIN & 0.95 & 0.79 & 0.86 & -- & -- \\
		ALL & 0.94 & 0.78 & 0.86 & 0.78 & 0.63 \\
		\midrule
		\textbf{Subtask C} & \textbf{Precision} & \textbf{Recall} & \textbf{F1 score} & \textbf{Accuracy} & \textbf{MacroF} \\
		\midrule
		IND & 0.91 & 0.62 & 0.74 & -- & -- \\
		GRP & 0.23 & 0.94 & 0.37 & -- & -- \\
		OTH & 0.00 & 0.00 & 0.00 & -- & -- \\
		ALL & 0.37 & 0.52 & 0.36 & 0.45 & 0.37 \\
		\bottomrule
	\end{tabular}
\end{table*}

\begin{table*}[ht!]
	\centering
	\caption{XLNet results on all subtasks (A, B, and C)}
	\label{tab:xlnet_all_subtasks}
	\begin{tabular}{lccccc}
		\toprule
		\textbf{Subtask A} & \textbf{Precision} & \textbf{Recall} & \textbf{F1 score} & \textbf{Accuracy} & \textbf{MacroF} \\
		\midrule
		NOT & 0.91 & 0.82 & 0.86 & -- & -- \\
		OFF & 0.62 & 0.78 & 0.69 & -- & -- \\
		ALL & 0.62 & 0.78 & 0.69 & 0.81 & 0.78 \\
		\midrule
		\textbf{Subtask B} & \textbf{Precision} & \textbf{Recall} & \textbf{F1 score} & \textbf{Accuracy} & \textbf{MacroF} \\
		\midrule
		UNT & 0.38 & 0.78 & 0.51 & -- & -- \\
		TIN & 0.97 & 0.84 & 0.90 & -- & -- \\
		ALL & 0.97 & 0.84 & 0.90 & 0.83 & 0.71 \\
		\midrule
		\textbf{Subtask C} & \textbf{Precision} & \textbf{Recall} & \textbf{F1 score} & \textbf{Accuracy} & \textbf{MacroF} \\
		\midrule
		IND & 0.53 & 1.00 & 0.69 & -- & -- \\
		GRP & 0.80 & 0.26 & 0.39 & -- & -- \\
		OTH & 0.00 & 0.00 & 0.00 & -- & -- \\
		ALL & 0.44 & 0.41 & 0.36 & 0.56 & 0.36 \\
		\bottomrule
	\end{tabular}
\end{table*}

\section{Conclusion and Future Work}
\label{sec:conclusion}
As online platforms continue to grow, so does the prevalence of offensive and abusive language. Detecting such content before it spreads has become increasingly important for fostering safer digital environments. In response, numerous studies have explored machine learning techniques to address this issue, with deep learning methods consistently outperforming traditional approaches in text classification tasks.

In this work, we proposed an offensive language detection system based on XLNet, a generalized autoregressive pretraining model that has achieved state-of-the-art results on multiple NLP benchmarks. Our study employed the OLID dataset from the SemEval 2019 challenge, which features a three-level hierarchical annotation scheme: (A) offense detection, (B) offense categorization, and (C) target identification.

We conducted extensive experiments to evaluate the performance of XLNet and benchmarked it against BERT, a widely adopted model known for its effectiveness in similar tasks. The results demonstrate that XLNet outperforms BERT in the first two subtasks—detecting offensive language and categorizing its type. However, BERT showed a slight advantage in identifying the target of the offense (Level C). Furthermore, our experiments highlight the benefits of resampling techniques, particularly oversampling, in mitigating class imbalance and enhancing classification performance.

For future work, we plan to explore advanced ensemble strategies that combine the strengths of both XLNet and BERT. Additionally, applying domain adaptation techniques and leveraging larger, more diverse datasets may further improve performance, especially on underrepresented classes such as "OTH" in target identification.

\bibliographystyle{IEEEtran}

\end{document}